\documentclass[10pt,twocolumn,letterpaper]{article}

\usepackage[pagenumbers]{cvpr}      
\usepackage{multirow}
\usepackage{booktabs}
\usepackage{caption}
\usepackage{CJKutf8}
\usepackage[ruled,vlined]{algorithm2e}
\definecolor{cvprblue}{rgb}{0.21,0.49,0.74}
\usepackage[pagebackref,breaklinks,colorlinks,allcolors=cvprblue]{hyperref}

\def\paperID{1773} 
\def\confName{CVPR}
\def\confYear{2026}

\title{A Sanity Check for Multi-In-Domain Face Forgery Detection in the Real World}

\author{
Jikang Cheng$^{1}$\thanks{Equal contribution}, Renye Yan$^{1*}$, Zhiyuan Yan$^{1}$, Yaozhong Gan$^{2}$, Xueyi Zhang$^{3}$,\\ Zhongyuan Wang$^{4}$, Wei Peng$^{5}$, Ling Liang$^{1}$\thanks{Corresponding author} \\[0.2cm]
$^{1}$Peking University, $^{2}$Nanjing University, $^{3}$The Chinese University of Hong Kong, Shenzhen, \\ $^{4}$Wuhan University, 
$^{5}$Stanford University 
}

\begin{document}
\begin{CJK*}{UTF8}{gbsn}
\maketitle
\begin{abstract}
Existing methods for deepfake detection aim to develop generalizable detectors. Although ``generalizable'' is the ultimate target once and for all, with limited training forgeries and domains, it appears idealistic to expect generalization that covers entirely unseen variations, especially given the diversity of real-world deepfakes. Therefore, introducing large-scale multi-domain data for training can be feasible and important for real-world applications.
However,  within such a multi-domain scenario, the differences between multiple domains, rather than the subtle real/fake distinctions, dominate the feature space. As a result, despite detectors being able to \textbf{relatively} separate real and fake within each domain (i.e., high AUC), they struggle with single-image real/fake judgments in domain-unspecified conditions (i.e., low ACC).
In this paper, we first define a new research paradigm named \textbf{Multi-In-Domain Face Forgery Detection (MID-FFD)}, which includes sufficient volumes of real-fake domains for training. Then, the detector should provide definitive real-fake judgments to the domain-unspecified inputs, which simulate the frame-by-frame independent detection scenario in the real world. 
Meanwhile, to address the domain-dominant issue, we propose a model-agnostic framework termed DevDet (\underline{Dev}eloper for \underline{Det}ector) to amplify real/fake differences and make them dominant in the feature space. DevDet consists of a Face Forgery Developer (FFDev) and a Dose-Adaptive detector Fine-Tuning strategy (DAFT). Experiments demonstrate our superiority in predicting real-fake under the MID-FFD scenario \textbf{while} maintaining original generalization ability to unseen data.
\end{abstract}
\section{Introduction}
\label{sec:intro}

\begin{figure}[t!]
    \centering
    \includegraphics[width=1\linewidth]{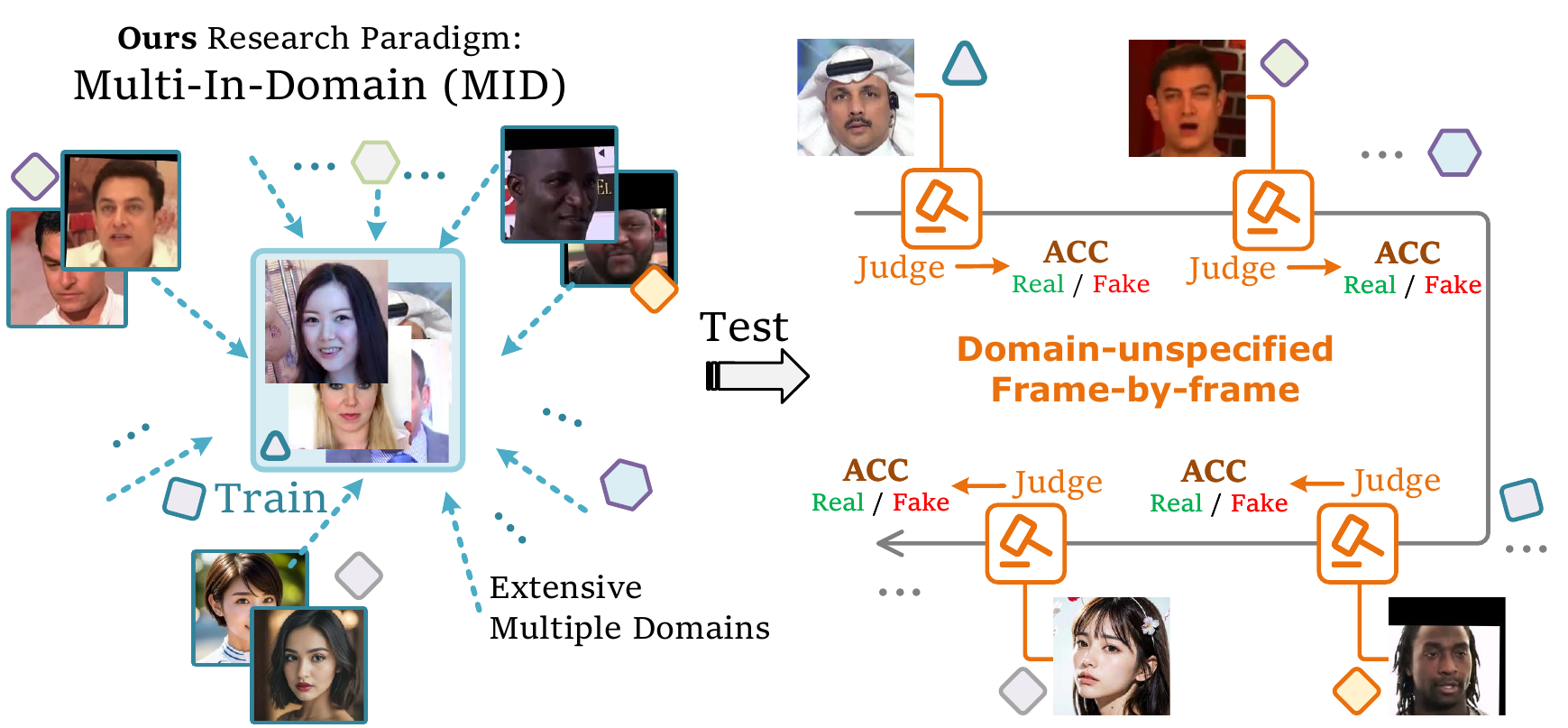}
    \caption{MID-FFD train on data with multiple domains and test on domain-unspecified inputs frame by frame with independent definitive real-fake judgment (\textit{i.e.}, ACC). Please refer to Fig.~\ref{fig:first-dist} for the challenge of MID-FFD.}
    \label{fig:first}
\end{figure}
\begin{figure*}[t!]
    \centering
    \includegraphics[width=1\linewidth]{figures/Intro_distribution.pdf}
    \caption{t-SNE visualization of detectors trained with two domains (D1: FF++~\cite{FF++}, D2: WDF~\cite{wilddeepfake}). Real and Fake within each specific domain are relatively well-divided, which is demonstrated by their promising in-domain AUC. However, in real-world applications,  the domain-unspecified test inputs cannot be directly judged as real or fake when they fall within the gap of the varied decision boundaries of D1 and D2, which is caused by the dominance of domain distinction over real/fake distinction in the feature space. Further visualization results could be found in Fig.~\ref{fig:exp_dist}. }
    \label{fig:first-dist}
\end{figure*}
The emergence of face forgery technologies presents serious societal risks, prompting growing concern among researchers. Consequently, the development of practical real-world detectors is critical for ensuring personal security and sustaining public confidence. Existing approaches~\cite{spsl,ucf,huang2023implicit,SBI,prodet,effort} mainly concentrate on training generalized face forgery detectors using limited (even single) data domains. 
Such a paradigm has already been criticized as overly idealistic~\cite{hdp,SURLID}, given that the increasingly diverse domains and characteristics of forgery data are unlikely to be comprehensively captured by extremely limited data domains, such as expecting generalization from outdated GAN-based face-swapping techniques to advanced SD-based entire face synthesis~\cite{df40}.
One popular attempt for multi-domain applications is introducing incremental face forgery detection (IFFD)~\cite{SURLID,dfil,hdp,iccv25,zhang2025choose}.
However, since face forgery detection (FFD) is a relatively simple binary classification task with inherently low training costs, the time savings achieved by IFFD are rather trivial compared to the catastrophic forgetting it suffered, particularly as the dataset scale increases.

In this paper, as illustrated in Fig.~\ref{fig:first}, we introduce a novel research paradigm termed multiple in-domain face forgery detection (MID-FFD), which we argue more accurately reflects the scenarios in the real world. The key question posed by MID-FFD is whether a detector can consistently deliver accurate and absolute real/fake judgments to frame-by-frame domain-unspecified inputs across multiple domains. 
Due to its conceptual simplicity, MID-FFD can be intuitively disregarded because of two widely accepted deductions. 
1) The strong performance achieved in the single-in-domain protocol can be seamlessly maintained when applied to multi-in-domain applications.
2) The relative discrimination of overall real/fake distribution within a single and same domain can be effectively transferred to the promising performance of absolute, frame-by-frame real/fake decisions in applications, where the inputs have unspecified and diverse domains.
However, in practice, these deductions are somehow misleading: although detectors may achieve coarse-level separation between real/fake distributions within each individual domain (\textit{i.e.}, higher AUC), they are limited in providing direct real/fake judgments for single images (\textit{i.e.}, lower ACC), under domain-unspecified conditions, as commonly encountered in real-world applications. As illustrated in Fig.~\ref{fig:first-dist}, both real and fake samples from  Domain1 are more closely aligned with real samples from Domain2, although they are both in-domain data. As a result, the domain-unspecified inputs that fall into the blue region cannot be reliably distinguished as either real or fake.
Such results reveal that \textbf{domain discrepancies may dominate over the subtle differences between real and fake}, thereby severely confusing the model's decision for direct real/fake judgments, which is crucial for real-world applications.

To address this challenge, we propose a two-stage framework termed DevDet that can amplify the distinctions between real and fake samples in a model-agnostic manner. By encouraging real-fake differences to dominate the learned distribution rather than being overshadowed by domain discrepancies, the detector can more confidently assess the authenticity of inputs from unspecified domains. Specifically, we first propose a Face Forgery Developer (FFDev), analogous to the \textit{photographic developer}, that is trained to expose potential forgery traces. FFDev is applied to the input prior to detection as a preprocessing step and is subsequently optimized via gradient-based feedback toward an improved confidence in identifying fake. FFDev is optimized using two types of samples: easy-real samples for real preservation and hard-fake samples for forgery exposure. Then, we propose a Dose-Adaptive Fine-Tuning (DAFT) strategy to help the pretrained detector to accommodate the developer-exposed images
Specifically, we fine-tune the pre-trained generalizable backbone using samples exposed by frozen optimized FFDev. While introducing FFDev improves the upper bound of detection performance since it enhances the detection confidence on hard samples, DAFT further introduce a Dose Dictionary (DoseDict) that can adaptively adjust the dose of the developer on an image-wise basis, thereby ensuring the lower bound of detection reliability and generalization ability. DoseDict is achieved by learning a dictionary that fits the hard samples from the training phase, and then determining the required developer dose for inference samples based on their reconstruction error with respect to the dictionary. 
The experiments demonstrate that our method enables more confident real/fake discrimination on the MID-FFD task, offering superior practical value compared to generalization and incremental learning approaches. Moreover, it can be applied to any existing pretrained generalizable backbone while fully preserving its original capability under extreme out-of-domain scenarios. Our contributions can be summarized as:
\begin{itemize}
    \item We propose to introduce the multi-in-domain face forgery detection (MID-FFD) task, which better reflects real-world deployment settings where data volumes and domain diversity are both large and extensive.
    \item By introducing the Face Forgery Developer (FFDev), we amplify potential forgery traces for any applied pretrained backbone, allowing real/fake differences to dominate over domain discrepancies.
    \item We propose a Dose-Adaptive Fine-Tuning (DAFT) strategy based on DoseDict to further improve the base detectors to adapt FFDev. It aims at both enhancing the MID effectiveness and fully maintaining the original capability of deployed pretrained detectors under extreme out-of-domain conditions.

\end{itemize}

\section{Related Work}\label{sec:relate}
\subsection{Generalizable Face Forgery Detection}
Current researchers mostly focus on the generalization of the detector to deal with the severe threat posed by face forgery. For example, given the observed model bias in the detector, various methods~\cite{ucf,Disentangle,ed4,fairNIPS} have been proposed to mitigate general model biases present in forgery samples. The advanced ViT-based methods like CLIP~\cite{clip} and the improved lora-based Effort~\cite{effort} are also proposed to enhance generalization ability. The model designs are also investigated in the latent space~\cite{prodet,lsda} and frequency space~\cite{spsl,F3Net}. In summary, many generalizable methods~\cite{ucf,prodet,lsda,ed4,F3Net,spsl,diffusionfake,minghui} are proposed to capture general forgery information from limited seen data and exhibit promising performance in a few unseen data. 
However, given the vast volume and diverse domains within existing forgery data, relying solely on a limited set of seen data to train an ideally generalizable detector is impractical.
\subsection{Development of Face Forgery Detection for Real-world Application}
The initial attempt to address AI-based malicious face abuse started with the effectiveness in the in-domain, such as the proposals of Xception~\cite{xception}, Capsule~\cite{capsule}, and MesoNet~\cite{mesonet}, which were introduced to enhance detection performance within a specified limited dataset. Subsequently, as deep models became increasingly powerful, and the FFD task is inherently a relatively simple binary classification problem—one that is easier to learn and fit—solutions for in-domain began to be considered well-established.
As a result, the research community shifted its focus toward the generalization problem, the current state of which and its shortcomings have been detailed above. To address the continuously evolving nature of forgery content, increasing researchers have begun to consider using the incremental learning paradigm to address deepfake detection in real-world scenarios. However, existing incremental learning methods~\cite{dfil,hdp,SURLID,iccv25} inevitably suffer from catastrophic forgetting. Additionally, although incremental training offers some efficiency advantages over multi-in-domain (MID) training, the FFD task is inherently efficient in terms of training and convergence. Therefore, these efficiency advantages appear minor in light of the performance drawbacks compared to data-centric MID-FFD. Moreover, they continue to adhere to a domain-by-domain evaluation strategy, indicating their insufficiency in addressing the frame-by-frame detection requirements of real-world scenarios. 

\begin{figure*}
    \centering
    \includegraphics[width=1\linewidth]{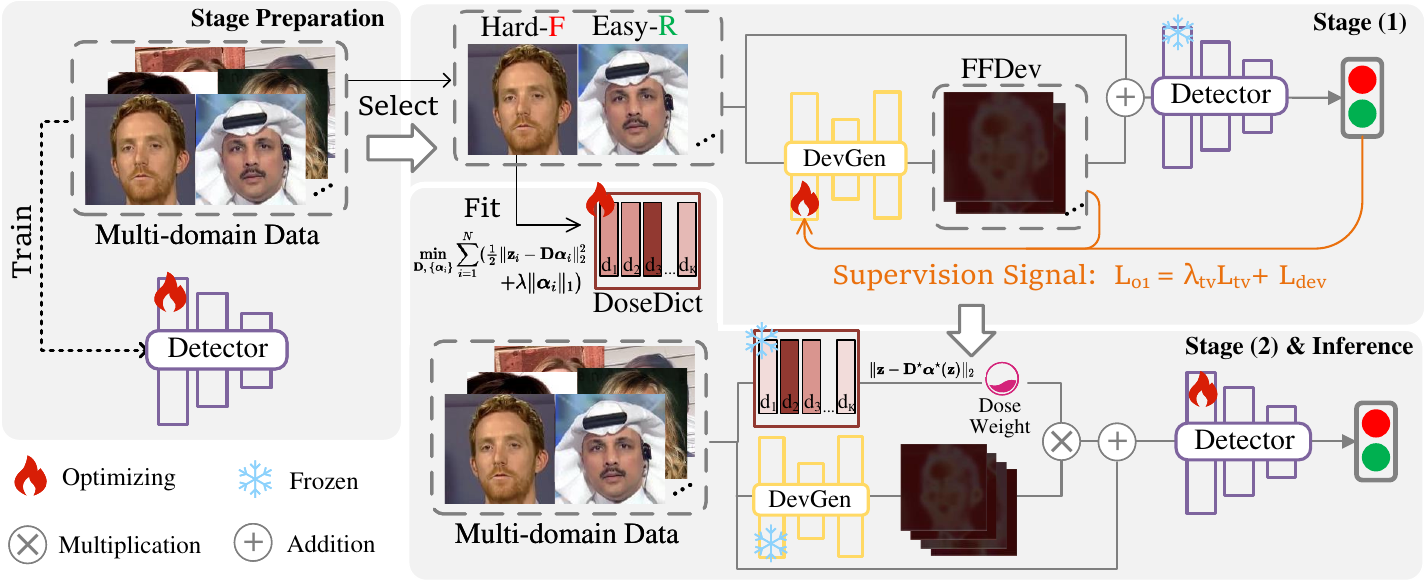}
    \caption{The two-stage architecture of the proposed DevDet.}
    \label{fig:main_arch}
\end{figure*}
\section{Motivation for MID-FFD}
Here, we concisely illustrate our motivation for introducing MID-FFD by answering two questions.\\
\textbf{Why can MID-FFD better reflect the real world?} Given the rapidly expanding diversity in both the real domains and intrinsic characteristics of forgery data, it is unrealistic to expect comprehensive representation using an extremely limited dataset. From a practical standpoint, existing generalization models in the research community primarily focus on sole training with the FaceForensics++~\cite{FF++} dataset, which contains only outdated graphic-based and early GAN-based forgeries from approximately five years ago, with limited domain diversity. Such data may include a small amount of real-domain information and GAN-specific artifacts, as well as some apparent blending traces. Intuitively, these cues could, in principle, be leveraged through careful method design to extract more general and subtle forgery features. However, they are evidently insufficient for detecting entirely out-of-domain forgeries, such as diffusion-based entire-face synthesis. Therefore, instead of relying solely on such limited data, it is more effective to expand the range of forgery types and real domains covered during training, and then ensuring accurate multi-in-domain performance on a large-scale dataset while maintaining potential generalization to out-of-domain scenarios.\\
\textbf{What is the challenge of MID-FFD?} Since real and fake can be effectively distinguished within in-domain settings, MID-FFD may initially seem to exhibit a similar property that could be easily achieved. However, once multiple domains are introduced into the learned latent space, the distinction between real and fake samples must be defined in absolute terms (typically with a confidence threshold of 0.5). In other word, while a model may achieve a high AUC in a specified domain, it can still perform poorly in terms of frame-by-frame real/fake accuracy, which is a more practical measure in real-world contexts. As illustrated in Fig.~\ref{fig:first-dist}, we observe that the inter-domain discrepancies surpass the real–fake discrepancies, which directly leads to the challenge of MID-FFD: Although the detector learns domain-specific relative real-fake differences, it performs limited on direct authenticity discrimination to inputs from the real world, where multiple domains are mixed and unspecified.   
\section{Method}\label{sec:method}
\subsection{Method Overview} Our method is a two-stage forgery developing method termed DevDet that could be deployed based on any pre-trained detectors. The core idea of DevDet is exposing the forgery traces, and thus enlarging the difference between real and fake. By doing so, we mitigate the influence of domain discrepancies and allowing authenticity-related variations to dominate the detector’s latent space, thus enabling effective binary real–fake discrimination for frame-by-frame, domain-unspecified inputs in real-world scenarios.  In Stage1, we optimize a Face Forgery Developer (FFDev) that post-process the inputs to exposure the forgery traces. In Stage2, we propose a Dose-Adaptive Fine-Tuning (DAFT) strategy to facilitate the adaptation of the pre-trained detector to images with FFDev. This is achieved by leveraging DoseDict, which can dynamically modulate the applied dose of FFDev. With respect to their functional roles, Stage1 enhances the detection confidence of fake inputs, Stage2 reschedules the latent space in response to the developed input, while DoseDict helps to maintain the original generalization ability of the base pre-trained detector. The overall pipeline of our method is shown in Fig.~\ref{fig:main_arch}.
\subsection{Preparation of data and model.} 
Prior to deploying our method, we first pre-train a detector using arbitrary architectures, such as Effnb4~\cite{effnet} or Effort~\cite{effort}. Specifically, the detector should be trained on a large multi-domain dataset $\mathcal{S}_m$ based on its official design, thus obtaining a pretrained MID detector  $f(\cdot,\theta_p)$, where $\theta_p$ is the pretrained weight. Then, based on the  detection confidence of the current detector, we can identify the hard fake (HF) samples and easy real (ER) samples in the training set, which can be formulated as:
\[
\mathcal{S}_{\text{HF}} = \{ \mathbf{x}_i \mid \text{Rank}(\text{Conf}(f(\mathbf{x}_i))) \in \text{Top-k}_{\text{low}}, \, \hat{y}_i = 1 \},
\]
\[
\mathcal{S}_{\text{ER}} = \{ \mathbf{x}_j \mid \text{Rank}(\text{Conf}(f(\mathbf{x}_j))) \in \text{Top-k}_{\text{low}}, \, \hat{y}_j = 0 \},
\]
where $\text{Rank}(\text{Conf}(f(\mathbf{x}_i)))$ refers to the ranking of the confidence score for each sample $\mathbf{x}_i \in \mathcal{S}_m$, $\text{Top-k}_{\text{low}}$ indicates the selection of the top $k$ samples with the scores closest to 0 (Real), and $\hat{y}_i$ is the ground-truth label of $\mathbf{x}_t$. Subsequently, the dataset $\mathcal{S}_1$ for training FFDev is constituted as $\mathcal{S}_1=\{\mathcal{S}_{\text{HF}},\mathcal{S}_{\text{ER}}\}$.
\subsection{Face Forgery Developer}
Similar to how a \textit{photo developer} transforms invisible film exposure into a visible photograph, Face Forgery Developer (FFDev) is designed to expose the forgery traces in fake data, thereby enlarging the real-fake difference for improved detection performance. Given an input image $\mathbf{x}\in \mathcal{S}_1$ and a Developer Generator (DevGen, $G(\cdot)$) based on an image reconstruction network~\cite{Generator}, we generate FFDev ($\boldsymbol{{\delta}}_\text{dev}$) as: 
\[ \boldsymbol{{\delta}}_\text{dev}=G(\mathbf{x},\theta_g) \in \mathbb{R}^{H\times W \times 3},
\]
where $\theta_g$ is the trainable parameter of $G$, $H$ and $W$ are the height and width of the input image. Subsequently, $\boldsymbol{{\delta}}_\text{dev}$ is added to $\textbf{x}$ as:
\begin{equation}
    \tilde{\mathbf{x}}=\mathbf{x}+\epsilon \boldsymbol{{\delta}}_\text{dev},\label{eq:apply}
\end{equation}
where $\tilde{\mathbf{x}}$ is the developed image, $\epsilon$ is the dose of $\boldsymbol{{\delta}}_\text{dev}$ that applied to the image. Subsequently, $\tilde{\mathbf{x}}$ will be predicted by the frozen $f(\cdot,\theta_p)$ as $y_p=f(\tilde{\mathbf{x}},\theta_p)$. The developing loss ($L_{dev}$) is a cross-entropy loss, which can be written as:
\begin{equation}
    L_{\text{dev}} = - \left( \hat{y} \log(y_p) + (1 - \hat{y}) \log(1 - y_p) \right).\label{eq:celoss}
\end{equation}
Notably, $L_\text{dev}$ contains two optimization objectives, that is, encouraging easy-real to maintain real while hard-fake to be predicted as fake. It is designed to firstly enforce consistency between the Easy-R sample and the sample with FFDev, ensuring that FFDev would not disrupt the original characteristics of real images. Meanwhile, $L_\text{dev}$ pushes the Hard-F, which was previously identified as real, to appear more fake, thereby amplifying the forgery characteristics and increasing the distinction between real and fake images.
Furthermore, we also introduce a Total Variation Loss ($L_\text{tv}$) to smooth FFDev as:
\[
L_\text{tv} = \sum_{i,j} \sqrt{(\tilde{\mathbf{x}}_{i+1,j} - \tilde{\mathbf{x}}_{i,j})^2 + (\tilde{\mathbf{x}}_{i,j+1} - \tilde{\mathbf{x}}_{i,j})^2},
\]
where $\tilde{\mathbf{x}}_{i,j}$ represents the pixel at $i$-th row, $j$-th line of image $\tilde{\mathbf{x}}$. $L_\text{tv}$ helps $\boldsymbol{{\delta}}_\text{dev}$ achieve better generalization and aids in convergence.

Therefore, the overall supervision signal for optimizing FFDev can be written as:
\begin{equation}
    L_\text{o1}=L_\text{dev}+\lambda_{tv}L_\text{tv},
\end{equation}
where $\lambda_{tv}$ is the trad-off parameter for $L_{tv}$. 
\subsection{Dose-Adaptive Fine-Tuning}
In Stage 2, we fine-tune the pretrained detector in order to re-organize the extracted feature space, enabling the differences between real and fake features after magnification to surpass domain discrepancies. This facilitates a more thorough adaptation to post-development images, ultimately enhancing the final MID detection performance. Additionally, we have developed the DoseDict, which dynamically adjusts the development dose based on the input's complexity. This further optimizes the entire forgery developer-based detection process, preserving both the MID-FFD performance enhancement and the original capability of the pretrained detector to generalize to out-of-domain data.
\subsubsection{DoseDict.}
DoseDict is a dictionary structure designed to learn and fit Hard Fake samples from MID data. It evaluates the fitness between input samples and the hard fake dictionary, and dynamically adjusts the dose of the developer based on this fit. It allows the model to apply a higher dose to the samples that are uncertain for decision-making, thus improving detection accuracy. On the other hand, when the input sample is judged to be simpler or falls outside the scope of MID knowledge, the FFDev dose is appropriately reduced, thereby maintaining the model's inherent generalization and detection performance.


\paragraph{Dictionary Training:}
To adaptively determine the appropriate developer dose based on the difficulty of the inference input, we conduct dictionary learning modeling.
Specifically, DoseDict can be written as $\mathbf{D} \in \mathbb{R}^{d \times K}$, where its column vectors are the dictionary atoms $\mathbf{d}_k$, $K$ is the dimension. The DoseDict training sample is $\mathbf{z}=f(\mathbf{x}^h)$, where $\mathbf{x}^h \in \mathcal{S}_{HF}$. Then, we train $\mathbf{D}$ through alternating training, where the overall training objective is:
\begin{equation}
\begin{aligned}
\min_{\mathbf{D},\,\{\boldsymbol{\alpha}_i\}} \quad
& \sum_{i=1}^{N}
\left(
    \tfrac{1}{2}\|\mathbf{z}_i - \mathbf{D}\boldsymbol{\alpha}_i\|_2^2
    + \lambda \|\boldsymbol{\alpha}_i\|_1
\right) \\
\text{s.t.} \quad
& \|\mathbf{d}_k\|_2 \le 1, \quad \forall k = 1,\dots,K,
\end{aligned}
\end{equation}
where $\boldsymbol{\alpha}$ is a sparse coding as the compressed representation of the input. We then perform \textbf{alternating} optimization: first optimizing $\boldsymbol{\alpha}$ with a frozen $\mathbf{D}$, followed by updating the $\mathbf{D}$ with a frozen $\boldsymbol{\alpha}$.\\
\textbf{When updating $\boldsymbol{\alpha}$}, we compute for each $\mathbf{z}_i$ as：
\begin{equation}
\boldsymbol{\alpha}_i^{(t)} 
= \arg\min_{\boldsymbol{\alpha}}
\left(
    \tfrac{1}{2}\left\|\mathbf{z}_i - \mathbf{D}^{(t-1)}\boldsymbol{\alpha}\right\|_2^2
    + \lambda \left\|\boldsymbol{\alpha}\right\|_1
\right).
\end{equation}
\textbf{When updating $\mathbf{D}$},
given $\mathbf{Z} = [\,\mathbf{z}_1,\ldots,\mathbf{z}_N\,] \in \mathbb{R}^{d\times N}$，
$\mathbf{A}^{(t)} = [\,\boldsymbol{\alpha}_1^{(t)},\ldots,\boldsymbol{\alpha}_N^{(t)}\,] \in \mathbb{R}^{K\times N}$. $\mathbf{D}$ is updated as
\begin{equation}
\mathbf{D}^{(t)} 
= \arg\min_{\mathbf{D}} 
\left\| \mathbf{Z} - \mathbf{D}\mathbf{A}^{(t)} \right\|_F^2,
\end{equation}
where $||\cdot||_F^2$ denotes Frobenius normalization. This process is iterated until convergence.

\paragraph{Dictionary Inference:}
During inference, we use the reconstruction error ($e(\mathbf{x})$) to measure the similarity between input $\mathbf{x}$ with hard fakes.
Formally, $e(\mathbf{x})$ is calculated by reconstructing with DoseDict:
\begin{equation}
    e(\mathbf{z}) = \|\mathbf{z} - \mathbf{D}^\star \boldsymbol{\alpha}^\star(\mathbf{z})\|_2.
\end{equation}

\subsubsection{Fine-Tuning with DoseDict}
Overall, given a training sample $\mathbf{x} \in \mathcal{S}_m$, its adaptive dose is $\epsilon_a=\text{Norm}(1-e(\mathbf{x}))$, and its corresponding FFDev can be written as $\delta_{dev}^h={G}(\mathbf{x}^h, \theta_g)$, where $Norm(\cdot)$ is a mapping to normalize dose value, and $\theta_g$ is frozen.
Subsequently, similar to Eq.~\ref{eq:apply}, we can then have $\tilde{\mathbf{x}}=\mathbf{x}+\epsilon_a\boldsymbol{{\delta}}_\text{dev}$ , and then $y_p=f(\tilde{\mathbf{x}},\theta_p)$, where $\theta_p$  is optimized by supervision same as Eq.~\ref{eq:celoss}.

The \textbf{inference process} is similar to the process of Stage2: 1) Obtain the adaptive dose $\epsilon_a$ for the unspecified input $\mathbf{x}$ using trained DoseDict. 2) Pre-process $\mathbf{x}$ with developer $\boldsymbol{\delta}_{\text{dev}}=G(\mathbf{x})$ as $\tilde{\mathbf{x}}=\mathbf{x}+\epsilon_a\boldsymbol{{\delta}}_\text{dev}$. 3) Predict the final real/fake confidence as $y_p=f(\tilde{\mathbf{x}})$.  


\begin{table*}[ht]
    \centering
    \caption{Performance comparison for Multi-In-Domain Face Forgery Detection based on Protocol 1.}
    \label{tab:main}
    \small
     \setlength{\tabcolsep}{5pt}      
    \begin{tabular}{lccccccccccc}
        \toprule
        \multirow{2}{*}{Method} 
        & \multirow{2}{*}{Venue}
        & \multicolumn{2}{c}{FF++~\cite{FF++}} 
        & \multicolumn{2}{c}{CDF~\cite{Celeb-df}} 
        & \multicolumn{2}{c}{DFDCP~\cite{DFDC-paper}} 
        & \multicolumn{2}{c}{WDF~\cite{wilddeepfake}} 
        & \multirow{2}{*}{S-AUC} 
        & \multirow{2}{*}{M-ACC} \\
        \cmidrule(lr){3-4}
        \cmidrule(lr){5-6}
        \cmidrule(lr){7-8}
        \cmidrule(lr){9-10}
        & & F-ACC & R-ACC & F-ACC & R-ACC & F-ACC & R-ACC & F-ACC & R-ACC & & \\
        \midrule
        Xception~\cite{xception} & CVPR'17 &  0.8732&  0.6797&  0.9655&  0.6016&  0.7362&  0.7797&  0.7981&  0.6097&  0.8431&  0.7555\\
        Capsule~\cite{capsule} & ICASSP’19 &  0.6672&  0.5918&  0.8102&  0.5575&  0.6393&  0.6631&  0.5917&  0.6732&  0.6854&  0.6493\\
        Effnb4~\cite{effnet} & ICML'19 &  0.9136&  0.6312&  0.9488&  0.5905&  0.8581&  0.7100&  0.5687&  0.8780&  0.8591&  0.7624\\
        F3Net~\cite{F3Net} & ECCV’20 &  0.8726&  0.6917&  0.9699&  0.5645&  0.8134&  0.6916&  0.5956&  0.8126&  0.8321&  0.7515\\
        CLIP~\cite{clip} & ICML’21 &  0.9012&  0.7179&  0.8931&  0.7201&  0.7932&  0.7117&  0.6045&  0.8237&  0.8810&  0.7707\\
        SPSL~\cite{spsl} & CVPR’21 &  0.9197&  0.6170&  0.9662&  0.6015&  0.8636&  0.7237&  0.5940&  0.8491&  0.8542&  0.7669\\
        SBI~\cite{SBI} & CVPR’22 &  0.8439&  0.9092&  0.7942&  0.7631&  0.6176&  0.7417&  0.5913&  0.6271&  0.7971&  0.7360\\
        IID~\cite{huang2023implicit} & CVPR’23 &  0.9012&  0.6831&  0.9610&  0.5616&  0.7065&  0.7396&  0.8052&  0.6314&  0.7869&  0.7487\\
        ProDet~\cite{prodet} & NeurIPS’24 &  0.8696&  \textbf{0.9124}&  0.8130&  0.7433&  0.6250&  0.7820&  0.6171&  0.7839&  0.8641&  0.7683\\
        Effort~\cite{effort} & ICML’25 &  0.9237&  0.7312&  0.9852&  0.5210&  0.8419&  0.7313&  0.6551&  \textbf{0.8821}&  0.8951&  0.7839\\
        \midrule
        \textbf{Ours} & — & \textbf{ 0.9317}&  0.8545&  \textbf{0.9856}&  \textbf{0.7671}&  \textbf{0.8690}&  \textbf{0.8978}& \textbf{ 0.8212}&  0.8701&  \textbf{0.9332}&  \textbf{0.8764}\\
        \bottomrule
    \end{tabular}
\end{table*}
\begin{table*}[ht]
    \centering
    \caption{Performance on a wider range of domains based on Protocol 2. The reported metric is average ACC on each dataset. }
    \label{tab:cross_dataset}
    \small
    \begin{tabular}{lccccccccccc}
        \toprule
        \multirow{2}{*}{Method} 
        & \multirow{2}{*}{FF++} 
        & \multirow{2}{*}{CDF} 
        & \multirow{2}{*}{DFDCP} 
        & \multirow{2}{*}{WDF} 
        & \multicolumn{4}{c}{DF40} 
        & \multicolumn{2}{c}{CDF3}  
        & \multirow{2}{*}{M-ACC} \\
        \cmidrule(lr){6-9}
        \cmidrule(lr){10-11}
        & & & & & BlendFace & SimSwap & DiT & SiT & AniTalker & FLOAT &  \\
        \midrule
        Xception &0.8379&0.8052&0.7808&  0.7613&  0.7643&  0.8715&  0.8082&  0.8270&  0.8056&  0.7679&    0.8029\\
        Effnb4 &0.8514&0.8172&0.7751&0.7039&0.8375& 0.8457&0.8134&0.8305&0.7912 &0.8015  &    0.8067\\
        CLIP &  0.8412&  0.8101&  0.7996&  0.6953&  0.7989&  0.8412&  0.8253&  0.8515&  0.8239&  0.7971&    0.8084\\
        SPSL &  0.8279&  0.8417&  0.8362&  0.7401&  0.7999&  0.8266&  0.8753&  0.8160&  0.7905&  0.8203&    0.8174\\
        Effort &  0.8757&  0.8675&  0.8513&  0.8012&  0.8736&  0.8810&  0.8659&  0.8432&  0.8099&   0.8171&  0.8486\\
        \midrule
        \textbf{Ours} &  \textbf{0.9270}&  \textbf{0.8971}&  \textbf{0.8852}&  \textbf{0.8601}&  \textbf{0.9071}&\textbf{0.9293}&\textbf{0.9376}&  \textbf{0.9401}&\textbf{0.8785}&  \textbf{0.8912}&  \textbf{0.9053}\\
        \bottomrule
    \end{tabular}
\end{table*}

\section{Experimental Results}\label{sec:exp}
\subsection{Setup}
\textbf{Datasets.} In this paper, a large scale of datasets is included for experiments, including Celeb-DF-v2 (CDF)~\cite{Celeb-df}, DeepFake Detection Challenge Preview (DFDCP)~\cite{dfdc}, FaceForensics++ (FF++)~\cite{FF++}, WildDeekfake (WDF)~\cite{wilddeepfake}, DiffusionFace (DiffFace)~\cite{diffusionface}, DF40~\cite{df40}, and Celeb-DF++ (CDF3)~\cite{cdfv3}, where CDF3, DiffFace, and DF40 incorporate multiple different advanced deepfake methods such as BlendFace~\cite{blendface}, Simswap~\cite{simswap}, DiT~\cite{dit}, SiT~\cite{sit}, AniTalker~\cite{anitalker}, FLOAT~\cite{float}, DDIM~\cite{ddim}, and DiffSwap~\cite{diffswap}. For training, we design two protocols to extensively evaluate MID-FFD performance. 
\begin{itemize}
    \item Protocol-1 (P1)=\{FF++, CDF) DFDCP, WDF\}: Classical datasets with distinct domains of both real and fake. 
    \item Protocol-2 (P2)=P1 + DF40 (\{SiT, DiT, BlendFace, SimSwap\}) + CDF3 (\{AniTalker, FLOAT\}): Larger-scale and more advanced datasets for a more faithful simulation of real-world MID-FFD scenarios.
\end{itemize}
\textbf{Baselines.} As this study constitutes the first evaluation of the MID-FFD task, no fully appropriate baseline methods currently exist. Consequently, we compare our approach with several representative generalizable models, including classical network backbones \{Xception~\cite{xception}, EffNet-B4~\cite{effnet}, Capsule~\cite{capsule}, CLIP~\cite{clip}\}, frequency-based methods \{F3Net~\cite{F3Net}, SPSL~\cite{spsl}\}, and designed strategies \{SBI~\cite{SBI}, IID~\cite{huang2023implicit}, ProDet~\cite{prodet}, Effort~\cite{effort}\}. All implementations are based on the reproductions available in DeepfakeBench~\cite{deepfakebench}. The landmarks of WDF is extracted via Dlib~\cite{king2009dlib} for ProDet and SBI, and the implicit identity constraint is ignored on WDF for IID.\\
\textbf{Metrics.} To properly investigate the scenarios of frame-by-frame detection with unspecified domains, we use fake accuracy (F-ACC) and real accuracy (R-ACC) to straightforwardly identify the real/fake classification capability. Moreover, to demonstrate the overall discrimination performance on the MID-FFD task, we adopt the Summarized AUC (S-AUC) metric. Instead of computing the AUC separately for each dataset, S-AUC summarizes all test sets into a unified evaluation benchmark, thereby preventing domain information leakage of the per-domain AUC.\\
\textbf{Implementation Details.} We strictly follow the official code and settings provided by the DeepFakeBench~\cite{deepfakebench} for face preprocessing.
Next, we carefully reproduce all baseline methods from DeepFakeBench, using the same training configuration to ensure a fair comparison. Ours is based on Effort for main comparison.
The Adam optimizer is applied with a learning rate of 0.0002, 10 epochs, an input size of 256 $\times$ 256 (224 for ViT-based models), and a batch size of 32. For stage1, we set the dose to the fixed $\epsilon=0.25$ to optimize the FFDev. Then, the $\epsilon_a$ has been also multiple with 0.25 in stage2 to align with stage1.
All experiments are performed on a single NVIDIA Tesla A100 GPU.
\subsection{Main Comparison}
Here, we present the performance of different methods on Protocol 1. It is first noteworthy that existing approaches are limited in achieving satisfactory detection performance under the MID-FFD scenario. As discussed in Sec. 3, although several prior studies have reported relatively favorable discriminative capability, the binary classification results (R-ACC / F-ACC) and the Summarized AUC remain highly problematic in domain-unspecified real-world detection settings.
In contrast, we consistently enhance the model’s confidence in binary discrimination, achieving up to a 11.80\% improvement in performance, thereby enabling a more reliable solution to the MID-FFD task. The result provides direct evidence of the effectiveness of our method.

Subsequently, we conducted further evaluations on a wider variety of datasets under Protocol 2 to more accurately simulate MID-FFD scenarios resembling real-world conditions. As the Real data in these datasets exhibit certain redundancies, the R-ACC metric unavoidably incorporates duplicate samples; therefore, we report only F-ACC, Mean F-ACC as M-ACC, and S-AUC. The results indicate that, due to the large volume of forgery data, the binary discrimination tasks exhibit substantially different performances. Nonetheless, our method consistently attains the highest overall performance. These findings highlight the strong potential of our approach for large-scale, real-world MID-FFD applications.

\begin{table}[t]
    \centering
    \caption{Model-agnostic enhancement of our method. ACC is reported for both MID-FFD enhancement and Cross-dataset maintenance compared to the base models. }
    \label{tab:model-ag}
    \footnotesize
         \setlength{\tabcolsep}{2.8pt}      
    \begin{tabular}{lcccc|ccc}
        \toprule
        \multirow{2}{*}{Methods} 
        & \multicolumn{4}{c|}{MID-FFD} 
        & \multicolumn{3}{c}{Cross-Dataset} \\
        \cmidrule(lr){2-5} \cmidrule(lr){6-8}
        & FF++ & CDF & DFDCP & WDF & DF40 & DiffFace &CDF3 \\
        \midrule
        Xception      &  0.7764&  0.7835&  0.7579&  0.7039&  0.7210&  0.6291&0.6996\\
        +Ours         &  0.8783&  0.8402&  0.8612&  0.8575&  0.7251& 0.6265& 0.6739\\
        \midrule
        Effnb4        &  0.7724&  0.7696&  0.7840&  0.7233&  0.7039&  0.6401&0.7039\\
        +Ours         &  0.8921&  0.8535&  0.8717&  0.8530&  0.7114& 0.6453& 0.7121\\
        \midrule
        SPSL          &  0.7683&  0.7838&  0.7936&  0.7215&  0.7693& 0.6216& 0.7316\\
        +Ours         &  0.8639&  0.8356&  0.8714&  0.8279&  0.7494& 0.6301& 0.7155\\
        \midrule
        Effort        &  0.8274&  0.7531&  0.7866&  0.7686&  0.8051& 0.6762& 0.7704\\
        +Ours         &  0.8931&  0.8763&  0.8834&  0.8456&  0.7935& 0.6869& 0.7767\\
        \bottomrule
    \end{tabular}
\end{table}
\subsection{Model-Agnostic Post-Processing}
As a model-agnostic method, we can enhance the MID-FFD performance of arbitrary pre-trained detectors while maintaining their original generalization capability. Therefore, in Tab.~\ref{tab:model-ag}, we implement our method to various mainstream detectors based on P1, and evaluate performance under both MID-FFD and Cross-Dataset (\textit{i.e.}, generalization) setting.\\
\textbf{MID-FFD Enhancement.} In Tab.~\ref{tab:model-ag} left, it can be observed that our method effectively optimizes the confidence of these models, resulting in more stable predictions and a substantial improvement in detection accuracy.\\
\textbf{Generalization Capability Preservation.} To demonstrate that the proposed method preserves the fundamental performance of the pre-trained detector, we conducted experiments to evaluate its generalization capability with or without our method. As shown in Tab.~\ref{tab:model-ag} right, our method effectively maintains the detector’s generalization performance while further enhancing its practical applicability.
\begin{table}[t]
    \centering
    \caption{Ablation study on the effectiveness of each proposed component (ACC). Cross represents the mean ACC among all cross-dataset evaluations.}
    \label{tab:ifdev_ablation}
    \footnotesize
    \setlength{\tabcolsep}{2.8pt}      
    \begin{tabular}{lccccc|c}
        \toprule
        Abl Variants & FF++ & CDF & DFDCP & WDF & M-ACC&Cross  \\
        \midrule
        Base           &  0.7724&  0.7696&  0.7840&  0.7233&  0.7624&  0.6826\\ \midrule
        +FFDev         &  0.8229&  0.8086&  0.8015&  0.6963&  0.7823&  0.5735\\
        +FFDev\&FixD  &  0.8742&  0.8433&  0.8529&  0.8401&  0.8526&  0.5851\\         +FFDev\&AdaD-P         &  0.8351&  0.8169&  0.8304&  0.7696&  0.8130&  0.6341\\\midrule
        +FFDev\&AdaD-S  &  0.8921&  0.8535&  0.8717&  0.8530&  0.8676&  0.6896\\
        \bottomrule
    \end{tabular}
\end{table}
\begin{figure*}[htbp]
    \centering
    \includegraphics[width=1\linewidth]{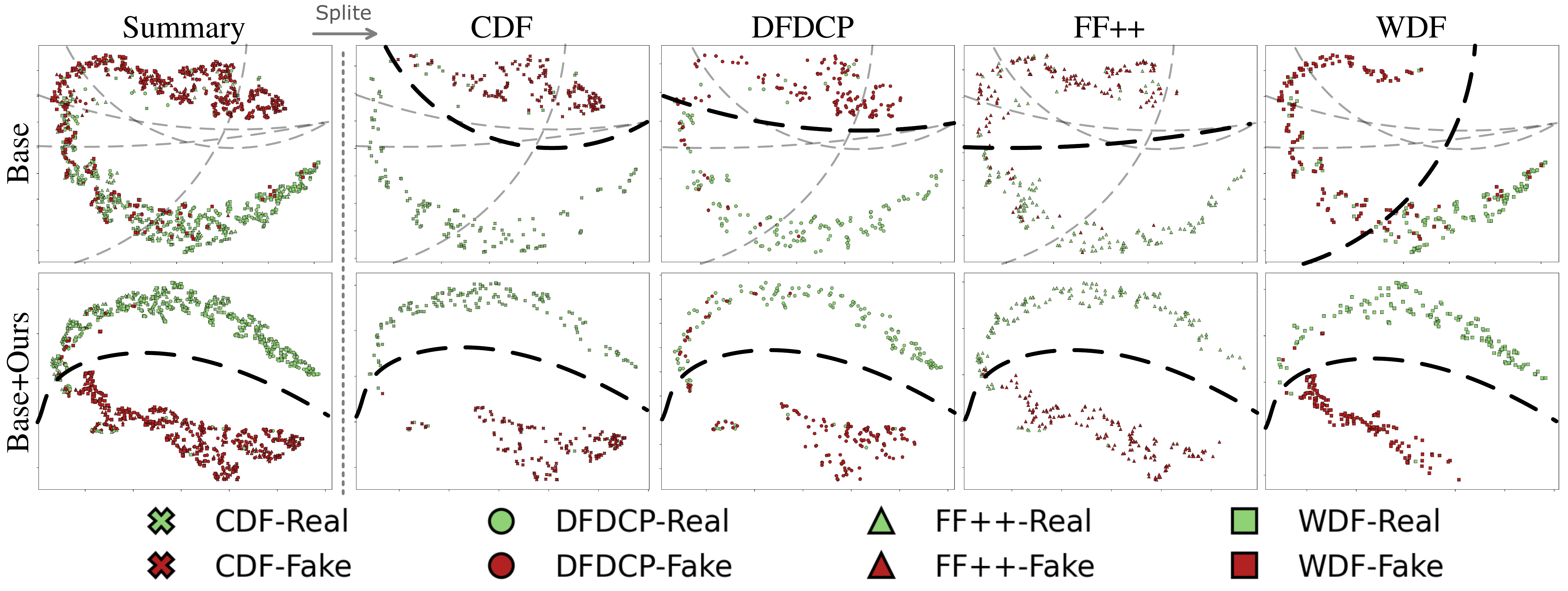}
    \caption{T-SNE~\cite{tsne} visualization of feature space. Here, Effnb4 is used as the base model that aligned with Tab.~\ref{tab:main}. The black dotted lines are the instruction lines of the possible decision boundary of each specified domain. MID Base has multiple distinct decision boundary across different domains, leading to the poor ACC when the input is domain-unspecified. Our result holds a consistent boundary for definitive real-fake judgment. Zoom in for better illustrations.}
    \label{fig:exp_dist}
\end{figure*}
\begin{figure}[htbp]
    \centering
    \includegraphics[width=1\linewidth]{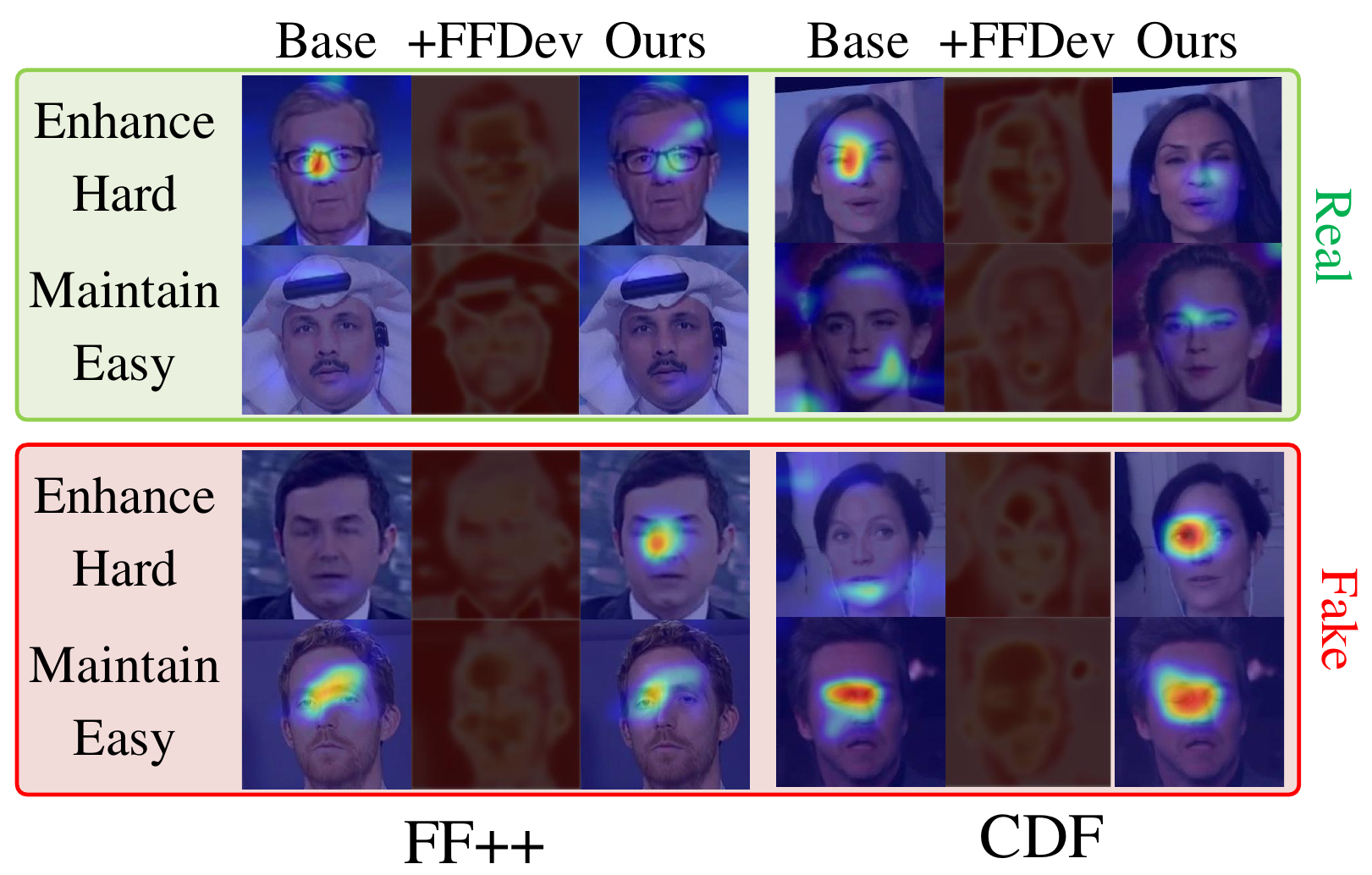}
    \caption{Grad-CAM~\cite{gradcam} visualization of the saliency map that is associated with classifying as \textit{fake}. We show two datasets and two conditions of Maintain Easy and Enhance Hard.}
    \label{fig:gradcam}
\end{figure}
\subsection{Ablation Study}
To investigate the specific impact of each proposed component, we designed the following ablation variables based on Effnb4 model: 1) Base: Base pre-trained Effnb4. 2) +FFDev: Introduce FFDev 3) +FFDev\&DFFT: Introduce FFDev with Dose-Fixed Fine-Tuning. 4) +FFDev\&DAFT-P: Parallelly optimize both FFDev and Detector via DAFT. 5) +FFDev\&DAFT-S: Sequentially train FFDev and then introduce DAFT for detector (Ours). It can be observed that directly incorporating FFDev enhances MID accuracy; however, this comes at the cost of a significant loss in the model's original generalization performance. Similarly, fine-tuning the model using a fixed-dose strategy does improve MID accuracy, but it still fails to preserve generalization performance.
Training the two stages in parallel can partially enhance MID performance while preserving the original generalization performance, despite neither task reaching its optimal result.
This may be because parallel training causes both FFDev and DAFT to perform limited in converging to their optimal states.
In contrast, using our two-stage approach with FFDev and adaptive-dose achieves the best improvement in MID detection while effectively maintaining generalization performance.

Furthermore, we also specifically analyzed the impact of the selection strategy and volume of the hard sample set, as well as the influence of parameters controlling the dose. Please refer to the \textit{Supplementary Material}.


\subsection{Visualization for Validation}
The visualized verification is crucial for understanding the existing challenge of vanilla MID detector and the rationale behind the superiorty of DevDet proposed in this paper. Here, we present results produced by two commonly used visualization strategies.\\
\textbf{Feature Visualization.} As shown in Fig.~\ref{fig:exp_dist}, we visualize feature space of Protocol 1 Effnb4 (Base) via t-SNE~\cite{tsne}. It can be clearly observed that the base model can coarsely divide real-fake within one specified domain, while it results in multiple decision boundaries for MID-FFD that will severely confuse the frame-by-frame domain-unspecified detection. In contrast, our method enhances the difference between real and fake, thus surpassing the dominant position of domain difference, thereby achieving promising MID detection performance. \\
\textbf{Saliency Map.} In Fig.~\ref{fig:gradcam}, we use Grad-CAM~\cite{gradcam} to visualize the saliency map of detected images with or without FFDev and DAFT.  We focus on the attention region that is associated with the model's decision for classifying the image as \textit{fake}. That is, the regions receiving attention are considered as containing evidence of manipulation. On two widely adopted datasets, we show the following advantages: 1) Fake-Enhance:
Our method can enhance the forgery that the base model fails to observe. 2) Real-Enhance: Introducing DAFT can more confidently distinguish images as Real if it has no forgery traces observed even with FFDev. 3) Maintain: Ours can maintain the original performance if the inputs are easy. These results further demonstrate the superior application potential of our method for MID-FFD.


\section{Conclusion}
In this paper, we analyze the existing generalizable-based methods regarding their insufficiency in forgeries and real domains during training. Then, we introduce the novel MID-FFD paradigm, which better reflects the application scenario of detecting forgery, but still struggles to achieve effective frame-by-frame domain-unspecified detection. To address this challenge, we propose DevDet that first introduce FFDev to expose the forgery traces, which amplifies the real-fake differences to dominate the latent space, and thus improves the MID performance. Then, we propose DAFT to maintain the original generalization ability of arbitrary backbones. Experiments demonstrate the superiority of our method.
{
    \small
    \bibliographystyle{ieeenat_fullname}
    \bibliography{refer,refer_incremental}

@InProceedings{SURLID,
    author    = {Cheng, Jikang and Yan, Zhiyuan and Zhang, Ying and Hao, Li and Ai, Jiaxin and Zou, Qin and Li, Chen and Wang, Zhongyuan},
    title     = {Stacking Brick by Brick: Aligned Feature Isolation for Incremental Face Forgery Detection},
    booktitle = {CVPR},
    month     = {June},
    year      = {2025},
    pages     = {13927-13936}
}

@article{prodet,
  title={Can We Leave Deepfake Data Behind in Training Deepfake Detector?},
  author={Cheng, Jikang and Yan, Zhiyuan and Zhang, Ying and Luo, Yuhao and Wang, Zhongyuan and Li, Chen},
  journal={NeurIPS},
  year={2024}
}

@article{ed4,
  title={ED $^{4}$: Explicit Data-level Debiasing for Deepfake Detection},
  author={Cheng, Jikang and Zhang, Ying and Zou, Qin and Yan, Zhiyuan and Liang, Chao and Wang, Zhongyuan and Li, Chen},
  journal={IEEE TIP},
  year={2024}
}

@String(IJCV = {Int. J. Comput. Vis.})

@String(CVPR= {IEEE Conf. Comput. Vis. Pattern Recog.})

@String(ICCV= {Int. Conf. Comput. Vis.})

@String(ECCV= {Eur. Conf. Comput. Vis.})

@String(NIPS= {Adv. Neural Inform. Process. Syst.})

@String(TIP  = {IEEE Trans. Image Process.})

@String(ACMMM= {ACM Int. Conf. Multimedia})

@String(ICASSP=	{ICASSP})

@String(IJCAI = {IJCAI})

@String(IJCV  = {IJCV})

@String(CVPR  = {CVPR})

@String(ICCV  = {ICCV})

@String(ECCV  = {ECCV})

@String(NIPS  = {NeurIPS})

@String(TIP   = {IEEE TIP})

@String(ACMMM = {ACM MM})

@String(ICML  = {ICML})

@inproceedings{spsl,
  title={Spatial-phase shallow learning: rethinking face forgery detection in frequency domain},
  author={Liu, Honggu and Li, Xiaodan and Zhou, Wenbo and Chen, Yuefeng and He, Yuan and Xue, Hui and Zhang, Weiming and Yu, Nenghai},
  booktitle=CVPR,
  pages={772--781},
  year={2021}
}

@article{ucf,
  title={UCF: Uncovering Common Features for Generalizable Deepfake Detection},
  author={Yan, Zhiyuan and Zhang, Yong and Fan, Yanbo and Wu, Baoyuan},
  journal=ICCV,
  year={2023}
}

@article{deepfakebench,
  title={Deepfakebench: A comprehensive benchmark of deepfake detection},
  author={Yan, Zhiyuan and Zhang, Yong and Yuan, Xinhang and Lyu, Siwei and Wu, Baoyuan},
  journal={arXiv preprint arXiv:2307.01426},
  year={2023}
}

@inproceedings{FF++,
  title={Faceforensics++: Learning to detect manipulated facial images},
  author={Rossler, Andreas and Cozzolino, Davide and Verdoliva, Luisa and Riess, Christian and Thies, Justus and Nie{\ss}ner, Matthias},
  booktitle=ICCV,
  pages={1--11},
  year={2019}
}

@inproceedings{xception,
  title={Xception: Deep learning with depthwise separable convolutions},
  author={Chollet, Fran{\c{c}}ois},
  booktitle=CVPR,
  pages={1251--1258},
  year={2017}
}

@Misc{dfdc,
note = {\url{https://www.kaggle.com/c/deepfake-detection-challenge} Accessed 2021-04-24},
author = {Deepfake detection challenge.}
}

@inproceedings{Celeb-df,
  title={Celeb-df: A new dataset for deepfake forensics},
  author={Li, Yuezun and Yang, Xin and Sun, Pu and Qi, Honggang and Lyu, Siwei},
  booktitle= CVPR,
  year = {2020}
}

@inproceedings{mesonet,
  title={Mesonet: a compact facial video forgery detection network},
  author={Afchar, Darius and Nozick, Vincent and Yamagishi, Junichi and Echizen, Isao},
  booktitle={IWIFS},
  pages={1--7},
  year={2018},
  organization={IEEE}
}

@inproceedings{Disentangle,
  title={Exploring Disentangled Content Information for Face Forgery Detection},
  author={Liang, Jiahao and Shi, Huafeng and Deng, Weihong},
  booktitle=ECCV,
  pages={128--145},
  year={2022},
  organization={Springer}
}

@inproceedings{effnet,
  title={Efficientnet: Rethinking model scaling for convolutional neural networks},
  author={Tan, Mingxing and Le, Quoc},
  booktitle=ICML,
  pages={6105--6114},
  year={2019},
  organization={PMLR}
}

@article{DFDC-paper,
  title={The deepfake detection challenge dataset},
  author={Dolhansky, Brian and Bitton, Joanna and Pflaum, Ben and Lu, Jikuo and Howes, Russ and Wang, Menglin and Ferrer, Cristian Canton},
  journal={arXiv preprint arXiv:2006.07397},
  year={2020}
}

@inproceedings{SBI,
  title={Detecting deepfakes with self-blended images},
  author={Shiohara, Kaede and Yamasaki, Toshihiko},
  booktitle=CVPR,
  pages={18720--18729},
  year={2022}
}

@inproceedings{huang2023implicit,
  title={Implicit Identity Driven Deepfake Face Swapping Detection},
  author={Huang, Baojin and Wang, Zhongyuan and Yang, Jifan and Ai, Jiaxin and Zou, Qin and Wang, Qian and Ye, Dengpan},
  booktitle=CVPR,
  pages={4490--4499},
  year={2023}
}

@inproceedings{F3Net,
  title={Thinking in frequency: Face forgery detection by mining frequency-aware clues},
  author={Qian, Yuyang and Yin, Guojun and Sheng, Lu and Chen, Zixuan and Shao, Jing},
  booktitle=ECCV,
  pages={86--103},
  year={2020},
  organization={Springer}
}

@inproceedings{capsule,
  title={Capsule-forensics: Using capsule networks to detect forged images and videos},
  author={Nguyen, Huy H and Yamagishi, Junichi and Echizen, Isao},
  booktitle=ICASSP,
  pages={2307--2311},
  year={2019},
  organization={IEEE}
}

@article{tsne,
  title={Visualizing data using t-SNE.},
  author={Van der Maaten, Laurens and Hinton, Geoffrey},
  journal={Journal of machine learning research},
  volume={9},
  number={11},
  year={2008}
}

@article{king2009dlib,
  title={Dlib-ml: A machine learning toolkit},
  author={King, Davis E},
  journal={JMLR},
  volume={10},
  pages={1755--1758},
  year={2009},
  publisher={JMLR. org}
}

@inproceedings{gradcam,
  title={Grad-cam: Visual explanations from deep networks via gradient-based localization},
  author={Selvaraju, Ramprasaath R and Cogswell, Michael and Das, Abhishek and Vedantam, Ramakrishna and Parikh, Devi and Batra, Dhruv},
  booktitle=ICCV,
  pages={618--626},
  year={2017}
}

@inproceedings{wilddeepfake,
  title={Wilddeepfake: A challenging real-world dataset for deepfake detection},
  author={Zi, Bojia and Chang, Minghao and Chen, Jingjing and Ma, Xingjun and Jiang, Yu-Gang},
  booktitle=ACMMM,
  pages={2382--2390},
  year={2020}
}

@article{df40,
  title={DF40: Toward Next-Generation Deepfake Detection},
  author={Yan, Zhiyuan and Yao, Taiping and Chen, Shen and Zhao, Yandan and Fu, Xinghe and Zhu, Junwei and Luo, Donghao and Yuan, Li and Wang, Chengjie and Ding, Shouhong and others},
  journal={arXiv preprint arXiv:2406.13495},
  year={2024}
}

@article{diffusionface,
  title={DiffusionFace: Towards a Comprehensive Dataset for Diffusion-Based Face Forgery Analysis},
  author={Chen, Zhongxi and Sun, Ke and Zhou, Ziyin and Lin, Xianming and Sun, Xiaoshuai and Cao, Liujuan and Ji, Rongrong},
  journal={arXiv preprint arXiv:2403.18471},
  year={2024}
}

@inproceedings{blendface,
  title={Blendface: Re-designing identity encoders for face-swapping},
  author={Shiohara, Kaede and Yang, Xingchao and Taketomi, Takafumi},
  booktitle=ICCV,
  pages={7634--7644},
  year={2023}
}

@inproceedings{diffswap,
  title={Diffswap: High-fidelity and controllable face swapping via 3d-aware masked diffusion},
  author={Zhao, Wenliang and Rao, Yongming and Shi, Weikang and Liu, Zuyan and Zhou, Jie and Lu, Jiwen},
  booktitle=CVPR,
  pages={8568--8577},
  year={2023}
}

@inproceedings{simswap,
  title={Simswap: An efficient framework for high fidelity face swapping},
  author={Chen, Renwang and Chen, Xuanhong and Ni, Bingbing and Ge, Yanhao},
  booktitle=ACMMM,
  pages={2003--2011},
  year={2020}
}

@inproceedings{lsda,
  title={Transcending forgery specificity with latent space augmentation for generalizable deepfake detection},
  author={Yan, Zhiyuan and Luo, Yuhao and Lyu, Siwei and Liu, Qingshan and Wu, Baoyuan},
  booktitle=CVPR,
  pages={8984--8994},
  year={2024}
}

@article{effort,
  title={Effort: Efficient orthogonal modeling for generalizable ai-generated image detection},
  author={Yan, Zhiyuan and Wang, Jiangming and Wang, Zhendong and Jin, Peng and Zhang, Ke-Yue and Chen, Shen and Yao, Taiping and Ding, Shouhong and Wu, Baoyuan and Yuan, Li},
  journal=ICML,
  year={2025}
}

@article{diffusionfake,
  title={Diffusionfake: Enhancing generalization in deepfake detection via guided stable diffusion},
  author={Chen, Shen and Yao, Taiping and Liu, Hong and Sun, Xiaoshuai and Ding, Shouhong and Ji, Rongrong and others},
  journal=NIPS,
  volume={37},
  pages={101474--101497},
  year={2024}
}

@inproceedings{dit,
  title={Scalable diffusion models with transformers},
  author={Peebles, William and Xie, Saining},
  booktitle=ICCV,
  pages={4195--4205},
  year={2023}
}

@inproceedings{sit,
  title={Sit: Exploring flow and diffusion-based generative models with scalable interpolant transformers},
  author={Ma, Nanye and Goldstein, Mark and Albergo, Michael S and Boffi, Nicholas M and Vanden-Eijnden, Eric and Xie, Saining},
  booktitle=ECCV,
  pages={23--40},
  year={2024},
  organization={Springer}
}

@article{ddim,
  title={Denoising diffusion implicit models},
  author={Song, Jiaming and Meng, Chenlin and Ermon, Stefano},
  journal={arXiv preprint arXiv:2010.02502},
  year={2020}
}

@inproceedings{clip,
  title={Learning transferable visual models from natural language supervision},
  author={Radford, Alec and Kim, Jong Wook and Hallacy, Chris and Ramesh, Aditya and Goh, Gabriel and Agarwal, Sandhini and Sastry, Girish and Askell, Amanda and Mishkin, Pamela and Clark, Jack and others},
  booktitle=ICML,
  pages={8748--8763},
  year={2021},
  organization={PmLR}
}

@article{cdfv3,
  title={Celeb-df++: A large-scale challenging video deepfake benchmark for generalizable forensics},
  author={Li, Yuezun and Zhu, Delong and Cui, Xinjie and Lyu, Siwei},
  journal={arXiv preprint arXiv:2507.18015},
  year={2025}
}

@article{Generator,
  title={Cross-domain transferability of adversarial perturbations},
  author={Naseer, Muhammad Muzammal and Khan, Salman H and Khan, Muhammad Haris and Shahbaz Khan, Fahad and Porikli, Fatih},
  journal=NIPS,
  volume={32},
  year={2019}
}

@inproceedings{anitalker,
  title={Anitalker: animate vivid and diverse talking faces through identity-decoupled facial motion encoding},
  author={Liu, Tao and Chen, Feilong and Fan, Shuai and Du, Chenpeng and Chen, Qi and Chen, Xie and Yu, Kai},
  booktitle=ACMMM,
  pages={6696--6705},
  year={2024}
}

@inproceedings{float,
  title={Float: Generative motion latent flow matching for audio-driven talking portrait},
  author={Ki, Taekyung and Min, Dongchan and Chae, Gyeongsu},
  booktitle=ICCV,
  pages={14699--14710},
  year={2025}
}

@inproceedings{fairNIPS,
  title={Fair Deepfake Detectors Can Generalize},
  author={Cheng, Harry and Liu, Ming-Hui and Guo, Yangyang and Wang, Tianyi and Nie, Liqiang and Kankanhalli, Mohan},
  year={2025},
  booktitle={NeurIPS},
}

@inproceedings{minghui,
  author= {Ming{-}Hui Liu and Harry Cheng and Tianyi Wang and Xin Luo and Xin{-}Shun Xu},
  title= {Learning Real Facial Concepts for Independent Deepfake Detection},
  booktitle= {IJCAI},
  pages= {1585--1593},
  year= {2025},
 }

@inproceedings{zhang2025choose,
  title={Choose Your Expert: Uncertainty-Guided Expert Selection for Continual Deepfake Detection},
  author={Zhang, Xueyi and Zhu, Peiyin and Sui, Jinping and Yang, Xiaoda and Tian, Jiahe and Lao, Mingrui and Cai, Siqi and Guo, Yanming and Tang, Jun},
  booktitle=ACMMM,
  pages={11502--11511},
  year={2025}
}

@inproceedings{dfil,
  title={Dfil: Deepfake incremental learning by exploiting domain-invariant forgery clues},
  author={Pan, Kun and Yin, Yifang and Wei, Yao and Lin, Feng and Ba, Zhongjie and Liu, Zhenguang and Wang, Zhibo and Cavallaro, Lorenzo and Ren, Kui},
  booktitle=ACMMM,
  pages={8035--8046},
  year={2023}
}

@article{hdp,
  title={Continual face forgery detection via historical distribution preserving},
  author={Sun, Ke and Chen, Shen and Yao, Taiping and Sun, Xiaoshuai and Ding, Shouhong and Ji, Rongrong},
  journal=IJCV,
  pages={1--18},
  year={2024},
  publisher={Springer}
}

@InProceedings{iccv25,
    author    = {Zhang, Xueyi and Zhu, Peiyin and Zhang, Chengwei and Yan, Zhiyuan and Cheng, Jikang and Lao, Mingrui and Cai, Siqi and Guo, Yanming},
    title     = {Generalization-Preserved Learning: Closing the Backdoor to Catastrophic Forgetting in Continual Deepfake Detection},
    booktitle = ICCV,
    month     = {October},
    year      = {2025},
    pages     = {3798-3808}
}
}

\clearpage
\setcounter{page}{1}
\maketitlesupplementary
\setcounter{section}{0}

\section{Details for Training and Evaluation}
\subsection{Training} As we mentioned, all results in this paper are reproduced based on the official code in DeepfakeBench~\cite{deepfakebench}. Specifically, the original data videos are sampled to 8 frames each for training and testing. The faces in each frame are detected and cropped via Dlib~\cite{king2009dlib}, and 10\% padding is maintained for each face image. During training, we introduce multiple data augmentations following the configuration of Effort~\cite{effort}, including HF (horizontal flip), BC (brightness–contrast adjustment), HSV (hue–saturation–value shift), IC (image compression), GN (Gaussian noise), MB (motion blur), CS (channel shuffle), CO (Cutout), RG (random gamma), and GB (glass blur). These augmentations are applied with preset probabilities to increase appearance diversity and improve the model’s robustness to illumination changes, noise, blur, compression artifacts, and partial occlusions. 
\subsection{Evaluation} 
For a binary classifier producing a continuous prediction score $s_i \in \mathbb{R}$, let
\[
\mathcal{P}=\{(s_i,y_i)\mid y_i=1\}, \qquad 
\mathcal{N}=\{(s_j,y_j)\mid y_j=0\}
\]
denote the sets of positive (``fake'') and negative (``real'') samples. \\
\textbf{AUC.} The AUC measures the probability that a randomly chosen positive sample is assigned a higher score than a randomly chosen negative sample:
\[
\mathrm{AUC}
= \frac{1}{|\mathcal{P}|\cdot|\mathcal{N}|}
\sum_{(s_i,1)\in\mathcal{P}}
\sum_{(s_j,0)\in\mathcal{N}}
\mathbb{I}(s_i > s_j)
+ \frac{1}{2}\mathbb{I}(s_i = s_j),
\]
where $\mathbb{I}(\cdot)$ is the indicator function.

\noindent
Given a decision threshold $\tau$, the predicted label is
\[
\hat{y}_i = 
\begin{cases}
1, & s_i \ge \tau, \\
0, & s_i < \tau.
\end{cases}
\]
\textbf{F-ACC}. The fake accuracy is defined as the fraction of positive (fake) samples correctly classified:
\[
\mathrm{Acc}_{\text{fake}}
= \frac{1}{|\mathcal{N}|}
\sum_{(s_j,0)\in\mathcal{N}}
\mathbb{I}(\hat{y}_j = 1).
\]
\textbf{R-ACC}. Similarly, the real accuracy measures the proportion of negative (real) samples correctly classified:
\[
\mathrm{Acc}_{\text{real}}
= \frac{1}{|\mathcal{P}|}
\sum_{(s_i,1)\in\mathcal{P}}
\mathbb{I}(\hat{y}_i = 0).
\]

\begin{algorithm}[h]
\caption{Deverloper for Detector (DevDet)}\label{alg:main}
		\KwIn{Dataset: $S_m = \{\mathbf{X}_{real},\mathbf{X}_{fake}\}$;
        Designed Detector $f(\cdot,\theta_p)$;
        Developer Generator $G(\cdot,\theta_g)$.
        }

        Initialize $f(\cdot)$ pretrained on $S_m$;

        Initialize dataset for Optimizing Developer

        $S_1 =\{S_{HF},S_{ER}\}$

        training stage 1 for developer
        
        \For{  $\mathbf{x} \sim S_1$}{

        predict developer $\delta_\text{dev}$ based on $\mathbf{x}$

        $\delta_\text{dev}=G(\mathbf{x},\theta_g)$

        apply developer to image

        $\tilde{\mathbf{x}}=\mathbf{x}+\epsilon \delta_\text{dev}$

        predict real/fake

        $y_p = f(\tilde{\mathbf{x}},\theta_p)$

        compute developer loss

        $L_\text{dev}=  - \left( \hat{y} \log(y_p) + (1 - \hat{y}) \log(1 - y_p) \right).$

        compute overall loss of stage 1

        $L_{o1}=L_\text{dev}+\lambda_{tv}L_\text{tv}$

        update $\theta_{g}$ based on $L_{o1}$ via backpropagation
        }

        prepare DoseDict $\mathbf{D}$

        $\min_{\mathbf{D},\,\{\boldsymbol{\alpha}_i\}} \quad \sum_{i=1}^{N}\left(
        \tfrac{1}{2}\|\mathbf{z}_i - \mathbf{D}\boldsymbol{\alpha}_i\|_2^2
        + \lambda \|\boldsymbol{\alpha}_i\|_1
        \right) $

        Dose Adaptive Fine-Tuning (DAFT) for $f(\cdot,\theta_p)$

        \For{  $\mathbf{x} \sim S_m$}{

            adaptively compute dose

            $\epsilon_a = \text{Norm}(1-||\mathbf{z}-\mathbf{D}^*\boldsymbol{\alpha}^*(\mathbf{z})||_2)$

            apply adaptive developer to image

            $\tilde{\mathbf{x}}=\mathbf{x}+\epsilon_a \delta_\text{dev}$
            
            predict real/fake

            $y_p = f(\tilde{\mathbf{x}},\theta_p)$

            compute DAFT loss

            $L_\text{daft}=  - \left( \hat{y} \log(y_p) + (1 - \hat{y}) \log(1 - y_p) \right).$

            update $\theta_p$ based on $L_{o1}$ via backpropagation
        }
        
		\KwOut{Trained $G(\cdot,\theta_g), f(\cdot)$, and $\mathbf{D}$.}  
\end{algorithm}
Based on these definitions, it could be clearly observed that AUC represents the relative division between real and fake samples. For example, supposing all fake samples are detected as $0.9$ while all real samples are detected as $0.8$, the AUC will be \textbf{100\%}. However, for real-world detection with a fixed accuracy threshold (maybe $\tau=0.5$), the above case will have an accuracy of 0.5, which is equal to random guess. Therefore, AUC cannot accurately measure the real-world application performance of deepfake detector, especially in MID-FFD scenario, where domain distinction surpasses the real-fake distinction, making the absolute real-fake decision even challenger.
\section{Algorithm}
The Algorithm of the DevDe is shown in Alg.~\ref{alg:main}. In the algorithm, we concisely illustrate the two stages of the proposed DevDet during training, including FFDev optimization, DoseDict fitting, and DAFT for the pretrained detector.

\section{Further Experiments}
\begin{table}[t]
    \centering
    \caption{Comparison across different sample selection strategies.}
    \label{tab:volume-setting}
    \footnotesize
    \setlength{\tabcolsep}{3pt} 
    \begin{tabular}{lcccccc}
        \toprule
        \multirow{2}{*}{Methods} 
        & \multirow{2}{*}{Volume} 
        & \multicolumn{4}{c}{Datasets} 
        & \multirow{2}{*}{Avg} \\
        \cmidrule(lr){3-6}
        & & FF++ & CDF & DFDCP & WDF &  \\
\midrule
Base & - &  0.7724 &  0.7696 &  0.7840 &  0.7233  &  0.7623\\
\midrule
\multirow{2}{*}{HF-only}
& Small & 0.7803 & 0.7762 & 0.7917 & 0.7351 & 0.7708\\
& Large & 0.8123 & 0.7975 & 0.8103 & 0.7372 & 0.7893\\
\midrule
\multirow{2}{*}{HF+HR}
& Small & 0.7831 & 0.7715 & 0.7931 & 0.7386 & 0.7716\\
& Large & 0.8144 & 0.8205 & 0.8153 & 0.7702 & 0.8051.\\
\midrule
\multirow{2}{*}{All}
& Small & 0.8181 & 0.8042 & 0.7980 & 0.7453 & 0.7914\\
& Large & 0.8495 & 0.8457 & 0.8593 & 0.8078 & 0.8405\\
\midrule
\multirow{2}{*}{HF+ER (Ours)}
& Small & 0.8921 & 0.8535 & 0.8717 & 0.8530 & 0.8675\\
& Large & 0.8950 & 0.8785& 0.8745 & 0.8755& 0.8809\\
\bottomrule
    \end{tabular}
\end{table}
\subsection{Effect of Sample Selection Strategy for Optimizing FFDev}
In this paper, we select Hard-Fake (HF) and Easy-Real (ER) to optimize the Face Forgery Developer. To demonstrate the effect of maintaining real while enhancing fake, we design the following ablation variants: 1) HF-only: Investigating the effectiveness of maintaining real. 2) HF and HR (Hard Real): Attempting to enhance both real and fake at the same time. 3) All: using unspecified training samples to retrain the FFDev.
All variants are considered with two versions, that is, the small set (5000 samples) and the large set (20000 samples).
Subsequent to these variants, the DAFT is conducted in the same way as usual. The experiments are conducted based on Effnb4 and protocol 1. As shown in Tab.~\ref{tab:volume-setting}, the first observation is that the results show limited sensitivity to the volume of training samples for the HF-only, HF+HR, and HF+ER models. This suggests that a small number of challenging samples are sufficient to effectively represent the forgery trace for training. Then, it can be observed that HF-only and HF+HR exhibit marginal improvement due to no Real sample as a relative reference for preservation. All performs better while still being inferior to HF+ER.

\begin{figure}
    \centering
    \includegraphics[width=1\linewidth]{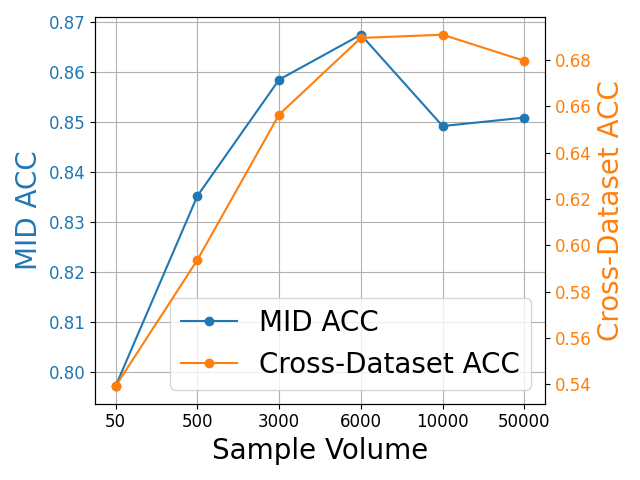}
    \caption{Effect of sample volume.}
    \label{fig:supp-vol}
\end{figure}
\subsection{Samples Volume for Training DoseDict}
As a sensitive hyperparameter, the number of samples to fit a DoseDict via dictionary learning is crucial for the accuracy and generalization ability of the predicted dose.  Here, we apply a range of sample numbers as an ablation study, which is shown in Fig.~\ref{fig:supp-vol}. It can be observed that, in the early stages of data augmentation, both generalization and detection accuracy show a certain degree of improvement. However, as the dataset grows too large, generalization performance gradually saturates, and the accuracy of MID experiences a slight decline. Therefore, this study selects 6000 as the optimal volume.
\end{CJK*}
\end{document}